%% file: main.tex
\documentclass{article}

\usepackage[preprint]{neurips_2026}

\usepackage[utf8]{inputenc} 
\usepackage[T1]{fontenc}    
\usepackage{hyperref}       
\usepackage{url}            
\usepackage{booktabs}       
\usepackage{amsfonts}       
\usepackage{nicefrac}       
\usepackage{microtype}      
\usepackage{xcolor}         
\usepackage{graphicx}
\usepackage[caption=false]{subfig}
\usepackage{wrapfig}
\usepackage{amsmath}
\usepackage{comment}
\usepackage{fvextra}

\newcommand\blfootnote[1]{%
  \begingroup
  \renewcommand\thefootnote{}\footnote{#1}%
  \addtocounter{footnote}{-1}%
  \endgroup
}

\title{Rate-distortion Assessment for Hierarchical Content Summarization}

\author{Yunye Gong$^1$, Indranil Sur$^1$ Jared Claypoole$^{1,\dagger}$, Pritish Sahu$^{1,\dagger}$ and Ajay Divakaran$^1$\\$^1$SRI International\\
\tt\small$^1${first.last@sri.com}}

\begin{document}

\maketitle
\blfootnote{$^{\dagger}$: work completed at SRI International}

\input{abstract}
\input{intro}
\input{rel_works}

\input{approach}

\input{exp}

\input{conclusion}

\bibliographystyle{plainnat}
\bibliography{custom}

\appendix
\section{Example Prompt for hierarchy construction}
\label{app:example-prompt}

\begin{Verbatim}[breaklines=true, breakanywhere=true]
**Long-Context Abstractive DAG**

You specialize in discourse planning for long documents (400–2,000+ words per chunk). Build an abstractive Directed Acyclic Graph (abs-DAG) that captures the hierarchy of ideas without copying text spans.

### Mindset
- Expect dense inputs; deliberately skim for a thesis before splitting.
- When unsure, err toward fewer, higher-level nodes and add depth only where the chunk truly diverges into distinct ideas.
- You may internally reference source sentences, but do **not** expose verbatim sub-documents in the final output.

### Procedure
1. **Global pass:** Identify the document’s primary thesis plus 2–6 major sections.
2. **Recursive planning:** For each node, decide whether it needs children. Require at least ~80–120 words of distinct content (or a clearly different rhetorical role) before you split.
3. **Summaries:** Every node summary must be a single abstractive sentence less or equal to 28 words that stands alone.
4. **Coherence checks:** Children must fully cover and collectively exhaust their parent’s scope. Do not overlap meanings between siblings.

### Output Format
Return a JSON list ordered top-down, breadth-first. Each entry must include:
```json
{
  "node_id": <int>,
  "parent_id": <int or null>,
  "summary": "<abstractive single-sentence summary>",
  "coverage_descriptor": "<brief note describing which portion of the document this node covers (e.g., 'intro + thesis', 'experiment details paragraphs 3-4')>",
  "children": [<child_id>, ...]
}
```
- `coverage_descriptor` is a short natural-language pointer (sentences, paragraph ranges, or thematic labels) that helps downstream aggregation when chunks are recombined.
- Omit any `sub_document` or raw quote fields.
- If a node is a leaf, set `children` to an empty list.
### Final Safety Checks
- Ensure `node_id` values are unique integers starting at 0.
- The root node (`parent_id = null`) must summarize the entire input chunk.
- Verify that the list, when viewed as edges `(parent_id -> node_id)`, contains no cycles.
- Validate that JSON parses without additional explanation text.
### Input Chunk
\end{Verbatim}
\end{document}

%% file: abstract.tex
\begin{abstract}
Powerful generative models including large language and vision models have demonstrated success across a wide range of content generation, editing and manipulation tasks via end-to-end prompt-based generation.
However, current assessment of the quality of generative content manipulation remains subjective, limiting scientific understanding of model capabilities. Inspired by rate-distortion theory from Shannon information compression theory, in this work we formalize hierarchical content summarization as an information encoding problem. We design a principled framework to extract the semantic hierarchy of the source content to support quantitative characterization of the rate-distortion trade-off between compression efficiency (rate) against information loss (distortion) during generative content manipulation, which allows us to benchmark various models with respect to performance conditional on rate budget of interest. We demonstrate the use of the framework with empirical experiments on content summarization for long-context question-answering with state-of-the-art large language models, offering new insights into their efficiency and comprehension capabilities. We further demonstrate the use of the semantic hierarchy as guidance for performing effective retrieval augmented generation, showing improved rate-distortion characteristics against current baselines.
\end{abstract}

%% file: intro.tex
\section{Introduction}
Modern multi-modal large language models (LLMs) have demonstrated transformative capabilities on a wide range of content manipulation and understanding tasks. Despite these advances, the evaluation of generated contents for tasks such as summarization remains an open challenge. Existing metrics, such as ROUGE and related overlap-based measures, primarily assess lexical similarity to the reference and fail to capture deeper semantic properties~\citep{rouge,bleu}. Recent efforts leverage LLMs as model-based evaluators and probe summarization quality via question answering (QA), enabling evaluation on factual consistency and/or alignment with human judgments~\citep{wang20asking,questeval,geval}. However, there is still a lack of a principled framework for assessing the quality of generative content manipulation, especially with respect to its deviation from the editorial priority and semantic hierarchy of the source content.

To address this gap, we approach LLM generation from the perspective of information theory~\citep{info,rd}, where the quality of summarization can be naturally viewed through the lens of a rate-distortion tradeoff. In compression theory, a rate-distortion function describes the performance bound for compression algorithms by characterizing the minimal representation cost (rate) needed to preserve essential information at a particular fidelity (distortion). In the context of LLM-based content summarization, considering a particular combination of model and prompt as a compressor, the theoretical framework enables us to characterize the efficiency of a given compressor in terms of the amount of information gain given additional context budget. Furthermore, given the editorial intent of the original content, different pieces of information may not weigh evenly. This motivates us to establish a proper fidelity measure taking into account the semantic hierarchy and priority. In other words, a proper assessment for comparing summarization algorithms should answer the question on whether the most salient information is retained first before including subsidiary details.

In particular, we propose a QA-based hierarchical framework to provide empirical quantification of information preservation. The hierarchy serves a road map to probe, discretize and quantify the information encoded in the source content. To construct a theoretically grounded hierarchy, we draw inspiration from rhetorical structure theory~\cite{rst} and category theory in mathematics~\citep{category}. For discretization of information, we first decompose the source content into a set of assertions, then discretize each assertion into a set of QA pairs, where the question can be answered only based on the information provided by the assertion. This allows us to provide an operational definition of the fidelity and to quantify the amount of information encoded in the original context and any manipulation (e.g., summarization) of the context. Given each source content, we extract the semantic hierarchy as a directed acyclic graph (DAG). Each node describes an assertion and the corresponding set of QAs. The edge connection is formed following the information flow in the source content such that each level in the hierarchy provides an abstraction of the material at a different granularity. 
This enables us to construct a canonical hierarchy following the rhetorical structure of the source content, without any domain priors on format of the hierarchy. The constructed hierarchy provides a representation of the information encoded in the source content in the semantic space and therefore the proposed framework is generally applicable to various source modalities. In this work, we demonstrate the use of the framework for text-based comprehension tasks. 

Having a hierarchy of QAs instead of an arbitrary set of QAs about the source content enables us to provide a comprehensive evaluation of content manipulation beyond merely examining the preservation of facts. It allows us to further perform graded inspection on information preservation following the rhetorical priority of the source content. The principled assessment framework provides an automatic pipeline to generate a benchmark for various content comprehension tasks and we demonstrate it with respect to the example task of content summarization. We inspect state-of-the-art LLM-based summarization solutions with respect to their rate–distortion spectrum. The alignment between empirical rate-distortion curves and theoretically desired rate-distortion properties provides grounded assessment beyond point-wise accuracy scores. We further demonstrate the use of the proposed semantic hierarchy as the guidance to improve RAG performance against both embedding-based and summarization-based baselines. Our results provide insights into designing models and content manipulation schemes that are both compact and information-preserving.

In summary, our core contributions include the following:
\begin{itemize}
\item
We present a novel theoretical framework for assessment of large models' generation from the perspective of information preservation by probing rate-distortion trade-offs.
\item
We propose a principled way to extract the semantic hierarchy from source content following its rhetorical priority, enabling systematic assessment, probing and manipulation of the content.
\item 
We perform rate-distortion assessment across state-of-the-art LLMs on the use case task of content summarization to demonstrate comprehensive analysis of model behavior beyond existing benchmarks.
\item 
We propose a novel RAG solution leveraging the theoretically-grounded hierarchy construction and demonstrate improved rate-distortion trade-off against state-of-the-art baselines.
\end{itemize}

%% file: rel_works.tex
\section{Related Works}
\paragraph{Rate-distortion tradeoff of large models}
In information compression theory, a rate–distortion function characterizes the optimal tradeoff between rate (number of bits to represent the source) and reconstruction fidelity given a distortion metric~\citep{rd}. In practice, it provides a theoretical guidance for designing compression solutions by defining what’s achievable in principle. In this work, we draw inspiration from classic rate-distortion theory and provide principled empirical analysis of LLM generation.
With increasing popularity of LLMs in diverse tasks, several recent works proposed to inspect the function and design of LLMs. In ~\citep{young25}, the author considered the trade-off between the representation cost of the LLMs with respect to number of weights and their performance and proposed optimization scheme for LLM quantization. Other works model the trade-off with respect to cost of prompting and generation performance~\citep{Nagle24, lee25}. In comparison to existing works discussing tradeoff with respect to rate discretization at the level of tokens or model sizes, in this work, we propose novel formulation modeling information discretization with respect to high-level semantics, enabling interpretable assessment aligned with human comprehension applicable to a wide range of tasks and application.
\paragraph{Semantic and rhetorical hierarchies} 
Many works seek to establish hierarchical structures in data to support effective comprehension and reasoning. Classic Rhetorical Structure Theory (RST)~\citep{rst} models documents as hierarchical discourse structures connected through rhetorical relations. Several efforts seek to organize semantics or concepts and their relations at broader scales via knowledge graphs~\citep{wordnet,conceptnet}. Recent work also seeks to formalize the semantic structure of a multi-modal document as a mathematical category by considering QA pairs as objects and defining morphisms between when one can answer the other.~\citep{category}. In comparison, in this work, we draw inspiration from classic theories and leverage instruction following capability of large models to construct a semantic hierarchy that follows information flow and rhetorical priority of the original content, providing abstraction at multiple levels to support hierarchical assessment of content manipulation and comprehension. The formulation is further supported by empirical experiments.
\paragraph{long context QA}
Current long-context QA solutions largely fall into two broad categories : 1) architecture improvement to support processing largely increased context windows via, e.g., adoption of sparse attention\citep{longformer,bigbird} or memory compression \citep{rae2020,chevalier2023} in transformer networks; and 2) compressing long-context at inference time via summarization or retrieval-augmented methods that selectively retrieve relevant partial contexts. Recent approaches further exploit hierarchical and structured retrieval. Sarthi et al. proposed RAPTOR~
\citep{raptor} which recursively clusters and summarizes documents into a retrieval tree. Edge et al., proposed GraphRAG~\citep{graphrag}, which organizes extracted entities and relationships into graph-based representations for answering questions requiring global understanding. In comparison to existing approaches, in this work, we demonstrate the use of a principled hierarchy which abstracts information at different levels following the original rhetorical intent of the content and enables efficient retrieval with largely improved rate-distortion trade-off.

%% file: approach.tex
\begin{figure}[t]
    \centering
    \includegraphics[width=\columnwidth]{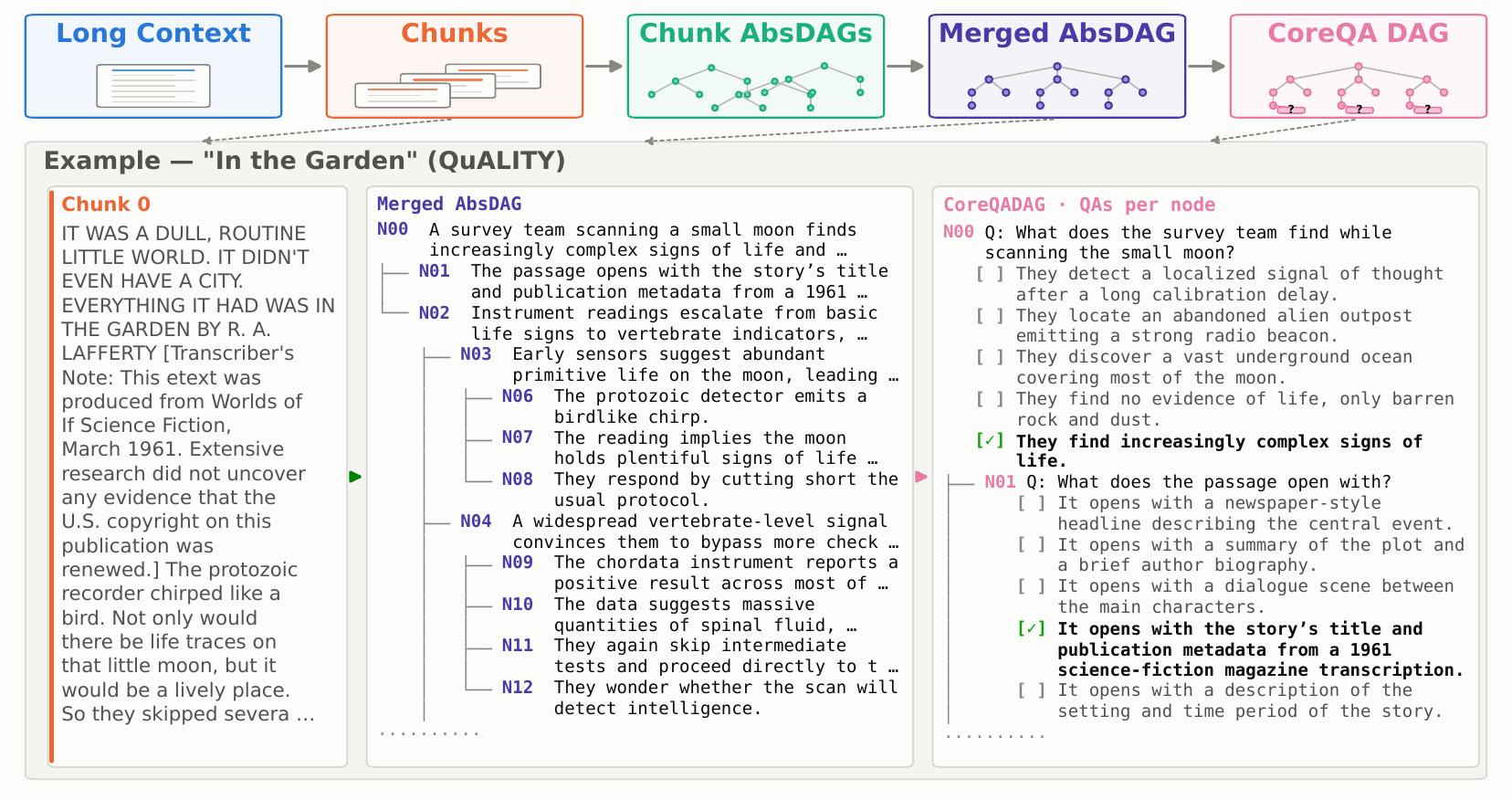}
    \caption{An overview of our proposed pipeline for extracting semantic hierarchy and decomposing information in QA pairs.}
    \label{fig:example}
\end{figure}

\section{Document Hierarchy Extraction}\label{sec:hierarchy}
To support systematic probing and characterization of information preservation given multi-modal inputs, we propose a novel framework modeling the information encoded in a source content via QA pairs. We consider the hierarchical set of questions that can be answered based on the source content and the corresponding answers as a representation of the encoded information. This formulation enables us to provide quantitative assessment of information preservation based on QA accuracy given variations of the original content. Conceptually, if two documents provide consistent answers to the same set of questions, we can consider the two documents to be equivalent from an information preservation perspective. This formulation further enables us to compare and differentiate documents both quantitatively and qualitatively. 
One key procedure to support the proposed formulation is to identify the proper set of QAs to serve as the operational checklist for examining a given document. This requires proper discretization and representation of encoded information. While in real-world document comprehension applications, specific sets of QAs may be provided as inputs based on the user interests, in this work, we propose a general solution without prior assumptions. Motivated by development in theoretically grounded hierarchy construction~\citep{rst,category}, we propose a discretization of the information taking into account the rhetorical structure and semantic priority in the original document and construct the hierarchal representation of the document as a directed acyclic graph. In particular, each node in the graph is represented by an abstractive assertion and corresponding QA pairs, while the edge in the graph relates two nodes such that QA on one node can be answered based on information embedded in the other. This construction enables us to represent the information at different resolutions as the child nodes always contain lower level details in comparison to the parent node. Operationally, we leverage LLM's instruction following to construct the proposed DAG. While theoretically the DAG hierarchy can be arbitrarily deep, in our empirical experiments, we rely on SOTA LLM's reasoning to consider only the first few levels in the hierarchical which corresponds to the set of key QAs describing the information in the document to a level of appropriate details. Figure~\ref{fig:example} shows an example of the proposed hierarchy construction to represent a long context as a DAG of QAs organized to reflect granular fidelity of comprehension.\par
Note that each of the QA pairs in the proposed structure can be easily translated into a corresponding assertion or claim as a natural representation of the information encoded in the original document. We consider the formulation based on QA to seamlessly support quantitative evaluation based on QA accuracy. \par
Also note that in comparison to efforts seeking to construct a universal hierarchy of semantics~\citep{wordnet,conceptnet}, our proposed effort provides a bottom-up yet general solution to construct proper hierarchy specific to each input document without domain assumptions on the document content or format. This formulation bypasses the effort of constructing an exhaustive semantic space with high dimensionality and focuses on the relevant semantics to the provided document. 

\section{Rate distortion analysis}
A key advantage of the proposed formulation is that it enables novel comprehensive evaluation of document comprehension via information theoretic metrics.
Specifically motivated by the rate-distortion analysis widely adopted as criteria to optimize compression design, we propose a set of assessments probing the operational rate-distortion tradeoff in LLM generation following the proposed hierarchy as a general road map.\par
In the context of document summarization, the proposed hierarchy provides, instead of a list of QAs with arbitrary order, a guidance to assess the information preservation with better alignment to the rhetorical intent in the original document. Let us consider a document represented by a hierarchy of QA pairs. For a certain level $m$ of the hierarchy, let there be $N$ pairs. We define a distortion criterion that is equivalent to Hamming Distortion $d(x,\hat{x})=
\begin{cases}
0, & \text{if } x = \hat{x},\\
1, & \text{if } x \neq \hat{x}.
\end{cases}$. Let there be a summary $S$ that consists of sentences that attempt to capture the essence of the meaning represented by QA pairs at level $m$. Suppose the summary helps answer $k$ out of the N questions. Then the distortion incurred by the summary $S$ is defined as $D(S)=N-k$ which is equivalent to cumulative Hamming distortion over all N questions if we consider the match of the summary with each of the question as a binary variable. Now that we have defined a distortion measure, the rate distortion function $R(D)$ where $D$ is the distortion defines the theoretical minimum rate required to achieve distortion $D$ as per
\begin{align}
    \inf_{Q_{Y|X}(y|x)}I_Q(Y;X) s.t. D_Q\leq D^*
\end{align} where $Q$ denotes the probability distribution of reconstructed information $Y$ given source information $X$ and $I$ denotes the mutual information between $X$ and $Y$.
The rate required to encode a QA pair is the minimum information required for answering the question which can be described by the corresponding assertion $A_i$. We define the rate $R(A_i)$ as the number of words required to state $A_i$. Note that alternative rate definitions can be adopted (e.g., the number of characters required to state $A_i$). As long as there is a consistent definition the rest of analysis will follow without loss of generality. Consider level $m$ with the set of asserstions $A={A_i}_{i=1,...,N}$, the rate distortion function $R(D)$ is then \begin{align}
\min_{s \in \binom{A}{k}} \sum_{a \in s} R(a)\end{align} computed by considering all possible subsets of assertions with size $k$ from collection of assertions $A$ with total size $N$.
Here we propose a pre-processing step. Let us order the rates $R_i$ from $1,...,N$ in an ascending order ${R_l, l=1,...,N}$, then the rate distortion function for distortion $N-k$ becomes $\sum_{l=1}^k R_l$. We thus have a precise definition of the rate distortion function at a given level $m$ of the hierarchy. A cumulative rate distortion function for the entire document can be constructed as a weighted sum of all $R(D)$ functions from each level of the hierarchy. Note that the relative weight of each level of the hierarchy is based on the task and is implicit. In the following analysis, we assume that at a given level $m$, all QA pairs are of equally weights corresponding to a source with uniform probability mass function.\par
Since we have a precise procedure for computing the rate distortion function for a document via projection on to a hierarchy of QA pairs, we can use the same distortion criterion for summaries that are generated through a chosen summarization algorithm and obtain its operational rate distortion curve. The ideal rate distortion curve gives us the lower bound and the distance from the ideal distortion curve characterizes the room for improvement in a practical summarization scheme. Also note that an ideal rate-distortion curve should possess monotonicity and convexity, an operational rate-distortion curve is not guaranteed to possess the properties as discussed Section~\ref{sec:results}. The alignment with the ideal properties provide insights for empirically determine the relative weights on information across different levels in the hierarchy.\par
This framework supports graded evaluation at different levels of abstraction. It enables us to compare different summarization strategies based on the same model as well as same summarization strategies based on different models. Instead of comparison based on a single accuracy metric, the assessment characterizes the trade-off in generation (given a particular combination of summarization strategy and model). Different curves can be compared such that gradient of the curve describes the efficiency of the generation strategy in terms of the amount of distortion reduction when more rate is allowed. Note that the proposed metric is a general analysis for context manipulation/prompt construction that can be transferred into a wide range of tasks with different operational definition for the rate-distortion trade-off. For example, for assessment of retrieval augmented generation (RAG) solutions, we can perform rate distortion analysis by measuring rate as the overall tokens used in RAG and measuring distortion based on the accuracy of RAG-based QA. For assessment of chain-of-thought strategies, we can again measure the rate based on the length or number of iterations in a chain-of-thought (CoT) construction and measure the distortion based on the accuracy of CoT-based QA.

%% file: exp.tex
\begin{figure*}[t]
\centering
\vspace{-1em}
\subfloat[]{\includegraphics[width=.5\textwidth]{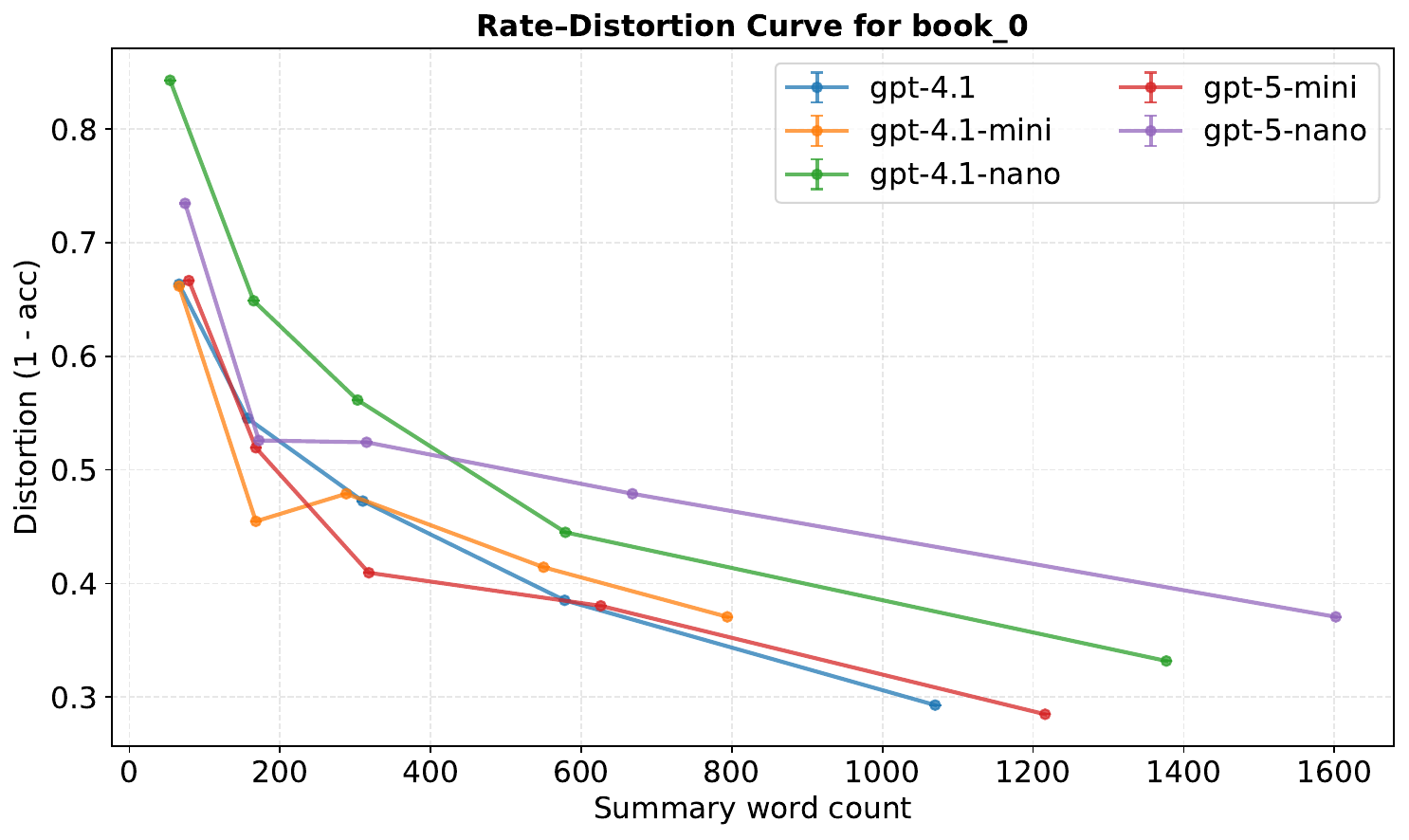}\label{fig:a}}
\hfill
\subfloat[]{\includegraphics[width=.5\textwidth]{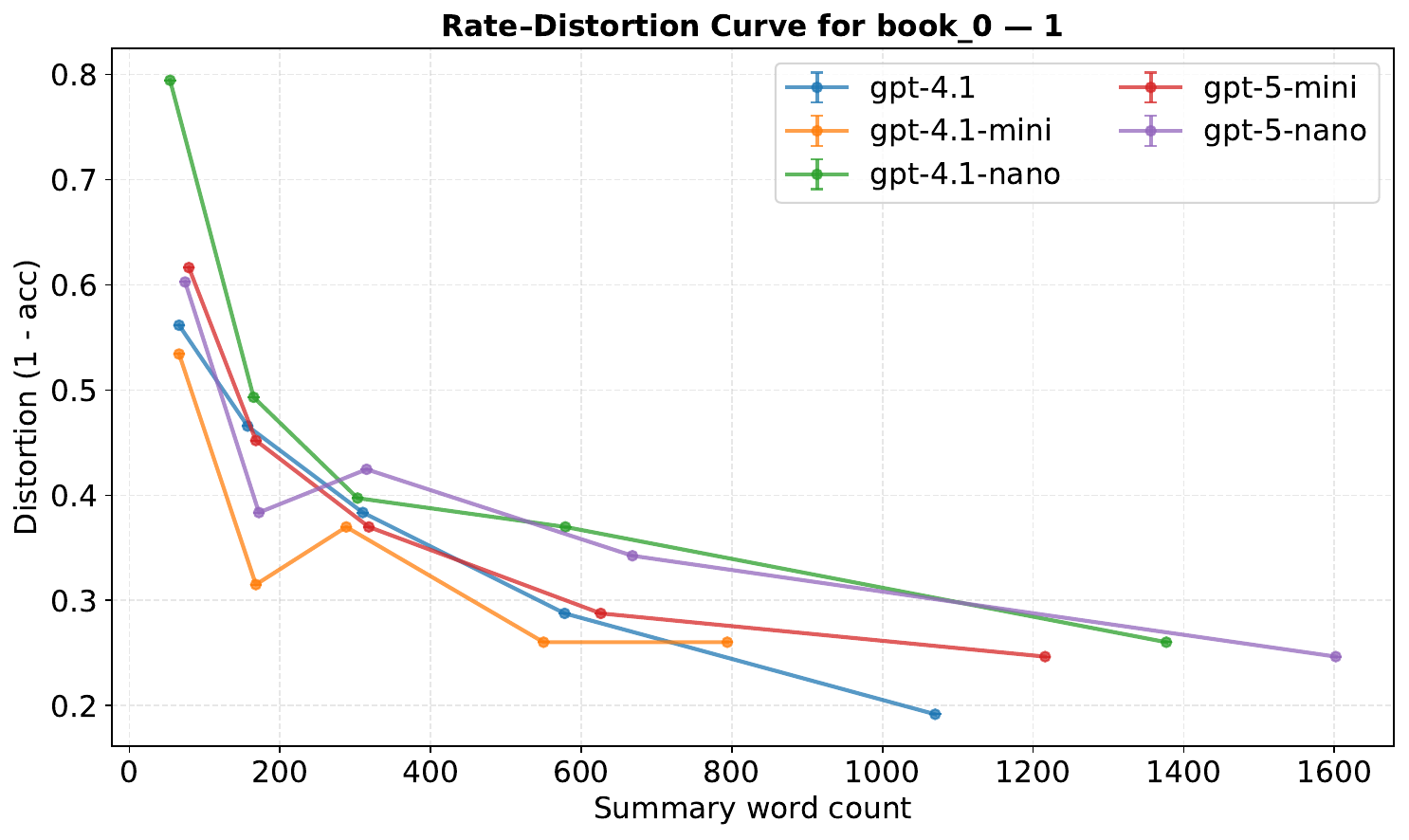}\label{fig:b}}
\hfill
\subfloat[]{\includegraphics[width=.5\textwidth]{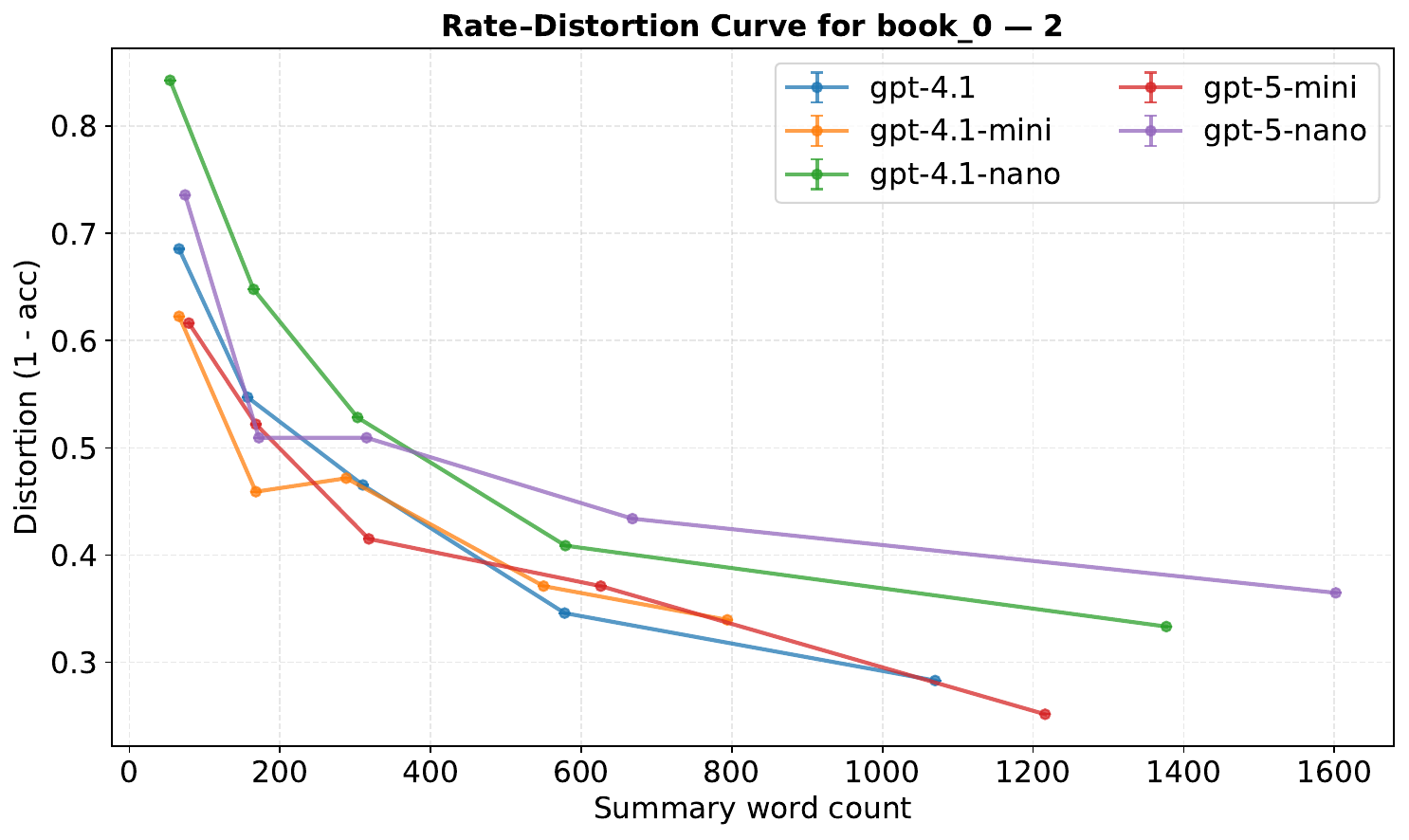}\label{fig:c}}
\hfill
\subfloat[]{\includegraphics[width=.5\textwidth]{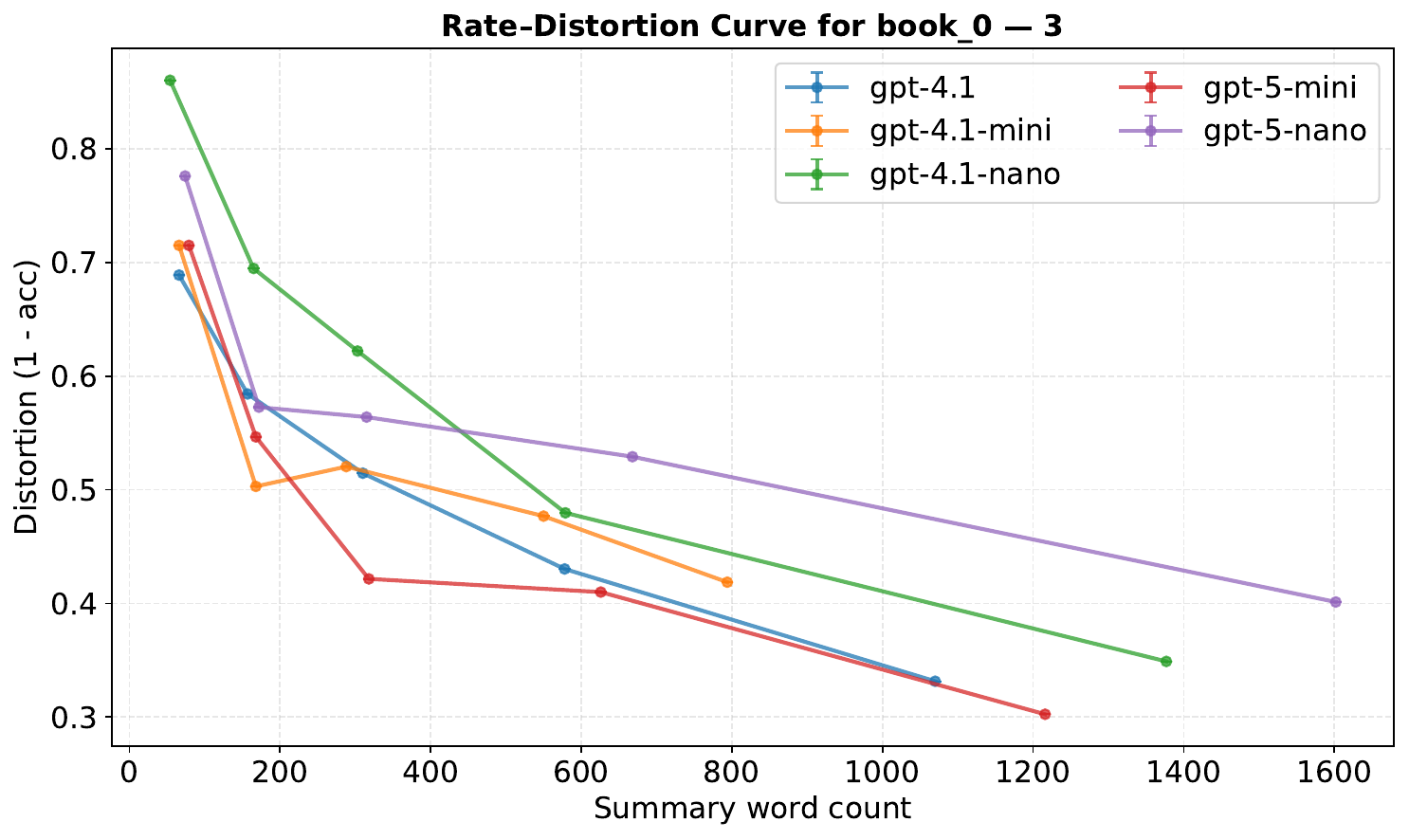}\label{fig:d}}
\caption{An example rate-distortion assessment of LLM content summarization on closed-source models. GPT-4.1-mini for QA evaluation. Panel (a) shows results on all QAs from the hierarchy. Panel (b)-(d) show results on QAs from higher to lower levels in the hierarchy respectively.}
    \label{fig:rd-close-1}
\end{figure*}

\begin{figure*}[h]
\centering
\vspace{-1em}
\subfloat[]{\includegraphics[width=.5\textwidth]{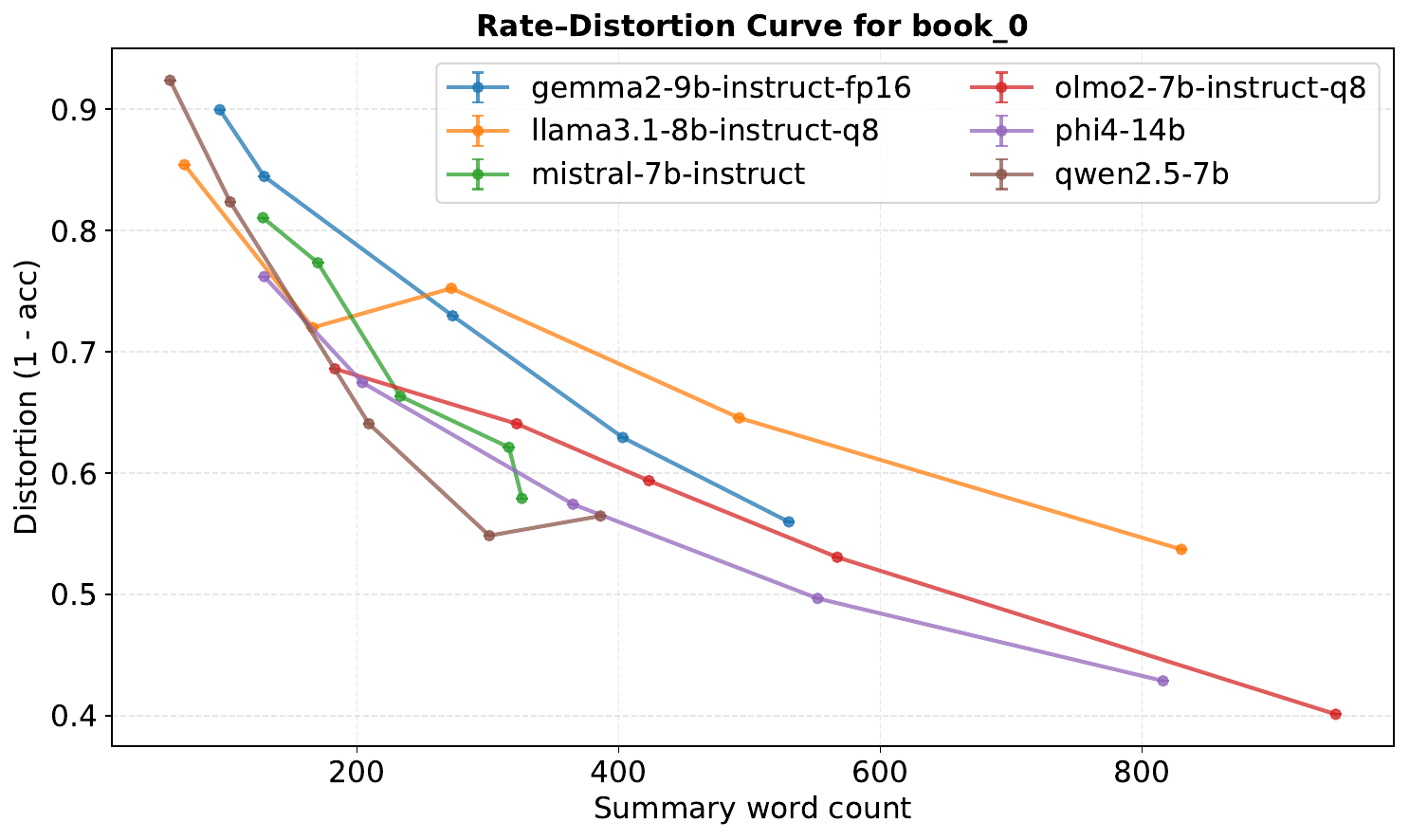}\label{fig:a}}
\hfill
\subfloat[]{\includegraphics[width=.5\textwidth]{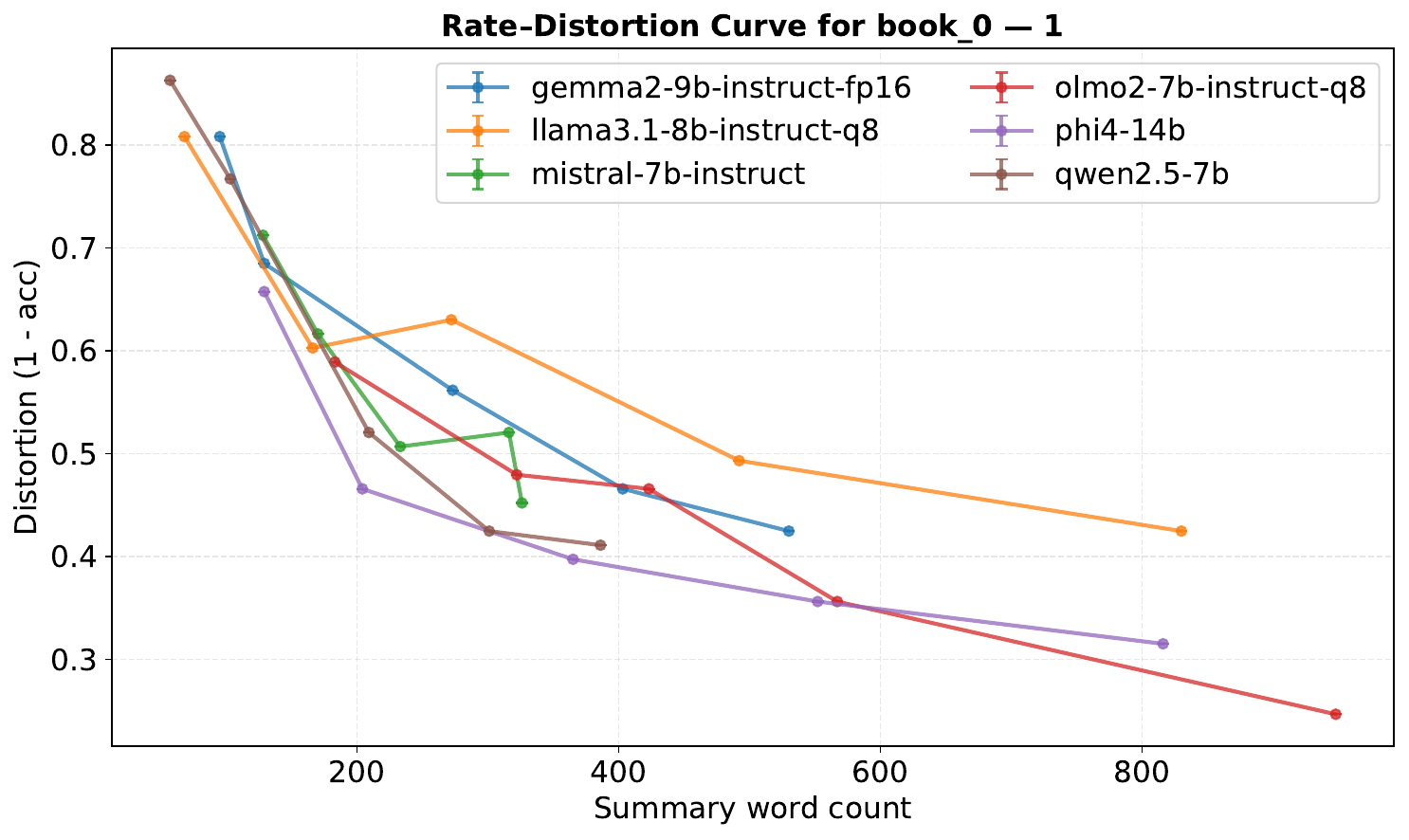}\label{fig:b}}
\hfill
\subfloat[]{\includegraphics[width=.5\textwidth]{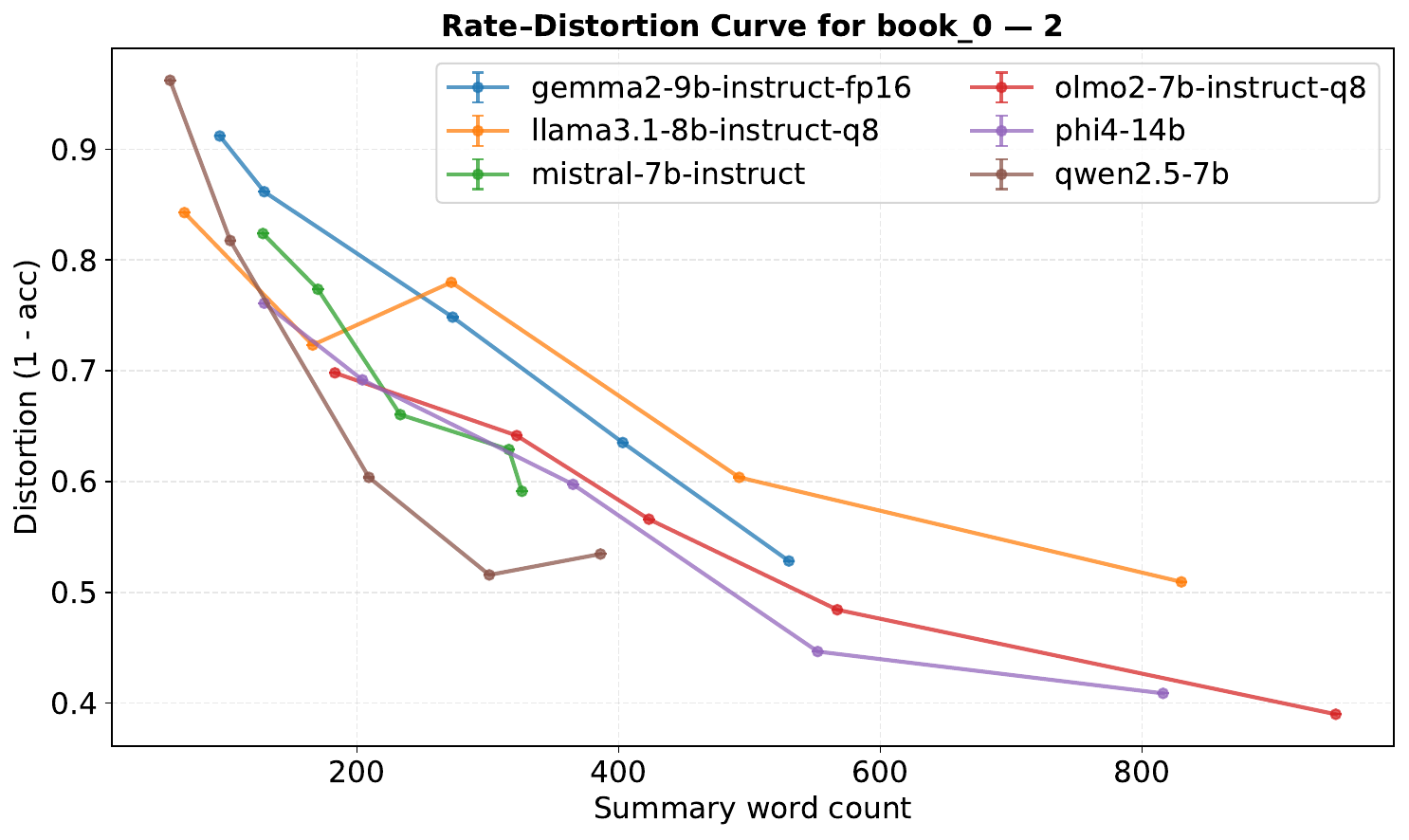}\label{fig:c}}
\hfill
\subfloat[]{\includegraphics[width=.5\textwidth]{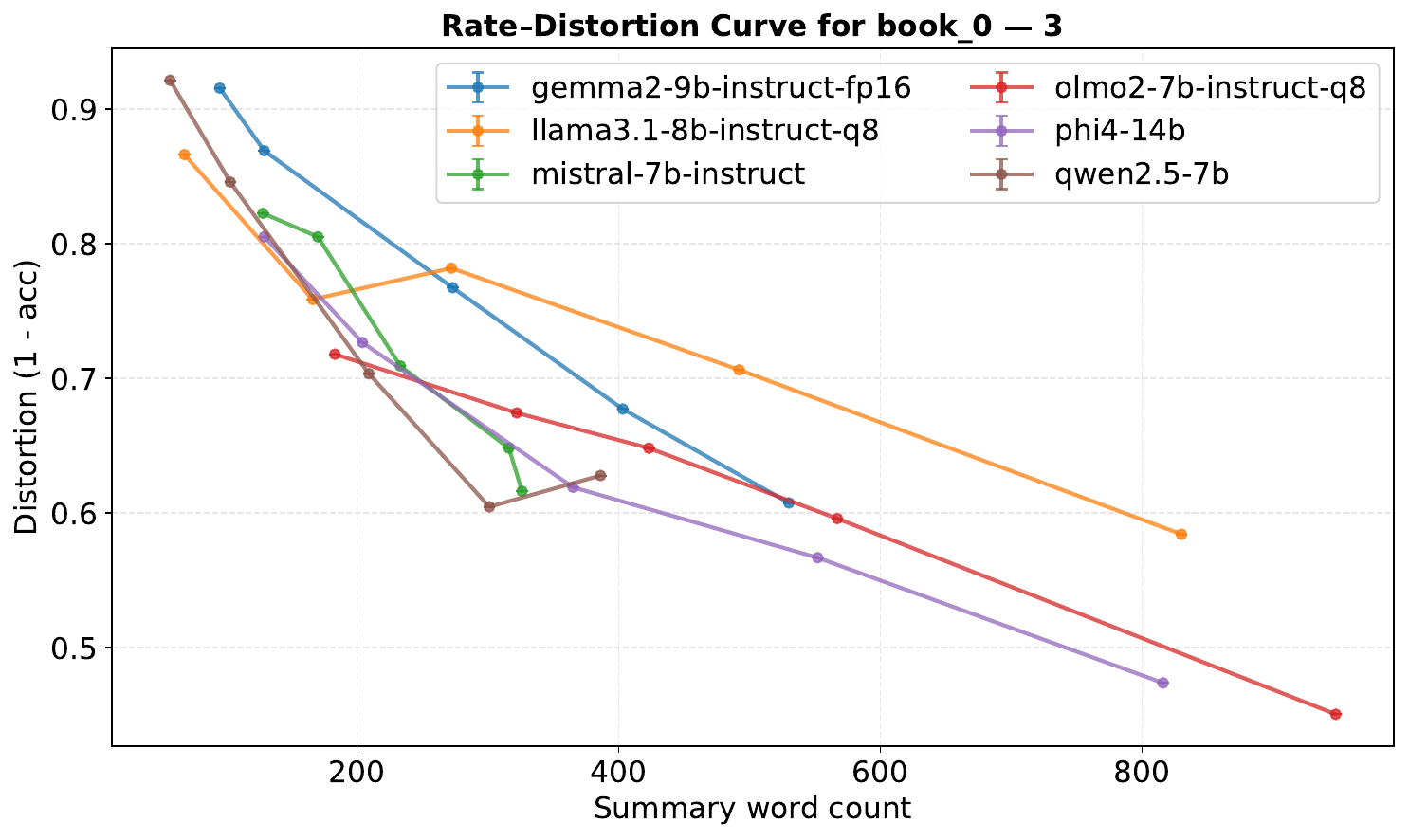}\label{fig:d}}
\caption{Rate-distortion assessment of LLM content summarization on open-source models. GPT-4.1-mini for QA evaluation. Panel (a) shows results on all QAs from the hierarchy. Panel (b)-(d) show results on QAs from higher to lower levels in the hierarchy respectively.}
    \label{fig:rd-open-1}
\end{figure*}

\section{Experimental Analysis}\label{sec:results}
To demonstrate the use of proposed hierarchy construction and rate distortion assessment, we perform a set of experiments with various SOTA LLMs.
\subsection{Data and Experimental Setting}
To demonstrate content summarization for long-context QA, we use the QuALITY\citep{quality} dataset which consists of multiple-choice QA samples with 300 long English context passages with an average length of around 5,000 tokens. There are in total 2523 QA. Data in QuALITY are manually constructed by annotators based on each entire context passage.\par
Given each context passage for QuALITY data sample, we construct the proposed semantic hierarchy as specified in Section~\ref{sec:hierarchy} using LLM instruction following. In comparison to QA samples from QuALITY dataset, this hierarchy provides a collection of QAs about the context passage at increasing level of granularity, enabling graded assessment grounded to the rhetoric priority and semantic content of the passage. Note that for construction of the hierarchy for each passage we do not impose a fixed structure of the hierarchy in terms of width and depth. Empirically we construct the hierarchy with arbitrary depth determined by the LLM generation. The resulting hierarchy can be considered as a truncated version of the theoretical hierarchy while focusing on higher-level semantics. Details on prompts adopted are included in the Appendix.\par
\textbf{Implementation Details:} The semantic hierarchy for each QuALITY passage is constructed by first chunking books into segments of at most 200 words with paragraphs as the delimiter, then processed through LLM prompted abstractive DAG summarization step that produces hierarchical abstractive summaries of the segment with per-node summaries and sub-documents (AbsDAG as shown in Figure~\ref{fig:example}). Once we have the AbsDAG of each chunk we merge them to create AbsDAG for the whole passage. This is followed by a LLM prompted QA generation step that decomposes each node summary and emits the level-annotated multiple-choice questions (CoreQA DAG as shown in Figure~\ref{fig:example}). We use GPT-5.2 to approximate the oracle hierarchy extraction process for the summarization assessments discussed in Section~\ref{sec:results_summary}. For the RAG experiments discussed in Section~\ref{sec:results_rag}, we used GPT-4.1-mini for the hierarchy construction considering the cost. 

\subsection{Rate-distortion assessment for summarization}\label{sec:results_summary}
We first demonstrate the proposed framework as a novel assessment scheme to provide principled and comprehensive examination of generative summarization. Intuitively, the capability to answer questions about a context passage at different levels provides a measure of the level of comprehension achieved on the material. Therefore, we measure the quality of a generated summary in terms of how well it preserves the key information in the original passage to support comprehension based on the summarized context. Given each context passage, we instruct the LLM to be evaluated to construct summaries at increasing lengths.

Figure~\ref{fig:rd-close-1} and Figure~\ref{fig:rd-open-1} demonstrate the proposed assessment on SOTA closed-source models and open-source models for summarization on example data from the QuALITY dataset. The horizontal axis measures the summary length and the vertical axis measures the distortion defined as $1-x$ where $x$ denotes the QA accuracy. In each figure, panel (a) shows the evaluation on the full set of QAs constructed in the semantic hierarchy. Panel (b), (c) and (d) in the figures show evaluation on QAs from higher to lower levels in the hierarchy respectively. While in this case, the evaluation on QA is also performed via LLM, we consider two experimental settings for ablation analysis for the quantitative assessment. Figure~\ref{fig:rd-close-1} shows the assessment of quality of summaries generated by different models while we use a fixed model as the oracle evaluator for QA based on the summaries. We used GPT-4.1-mini to balance reasoning capability and model bias. 
In all the experiments, we use the same GPT-5.2 model to construct the hierarchy of QAs.
In comparison to a single instance of QA evaluation with accuracy score, the proposed framework compares the performance of different models with respect to 2-D rate-distortion trade-offs which provides additional insights. For example, note that not all empirical rate-distortion curves satisfy the desired monotonicity. This implies the misbehavior of the model (e.g., orange curve in panel (a) of Figure~\ref{fig:rd-close-1}) with respect to efficiency as extra rate budget was not properly utilized in generation to achieve information reconstruction. One potential cause of the reduced accuracy with longer generation can be the existence of inconsistency or hallucination in the summaries which causes confusion of the QA evaluator.
In general, as shown in Figure~\ref{fig:rd-close-1}, we observe that the larger models tend to provide better trade-off shown as empirical rate-distortion curves with steeper gradient. The results also enable analysis over different ranges of the rate budgets. For example, as shown in panel (a) figure~\ref{fig:rd-close-1}, smaller models such as gpt-5-nano (purple) and gpt-4.1-mini (orange) are showing comparable performance to larger models in the regime of more aggressive summarization (e.g., rate less than 300 words) while the model finds it more challenging to utilize increasing rate budget effectively with persistent distortion levels (e.g., rate larger than 500 words). It is also observed that the smaller models tend to fall short in instruction following especially in generating summaries with longer target lengths (e.g., as shown in the orange curve in Figure~\ref{fig:rd-close-1}).

\begin{wrapfigure}{R}{0.5\textwidth} 
  \centering
  \vspace{-2em}
  \includegraphics[width=0.5\textwidth]{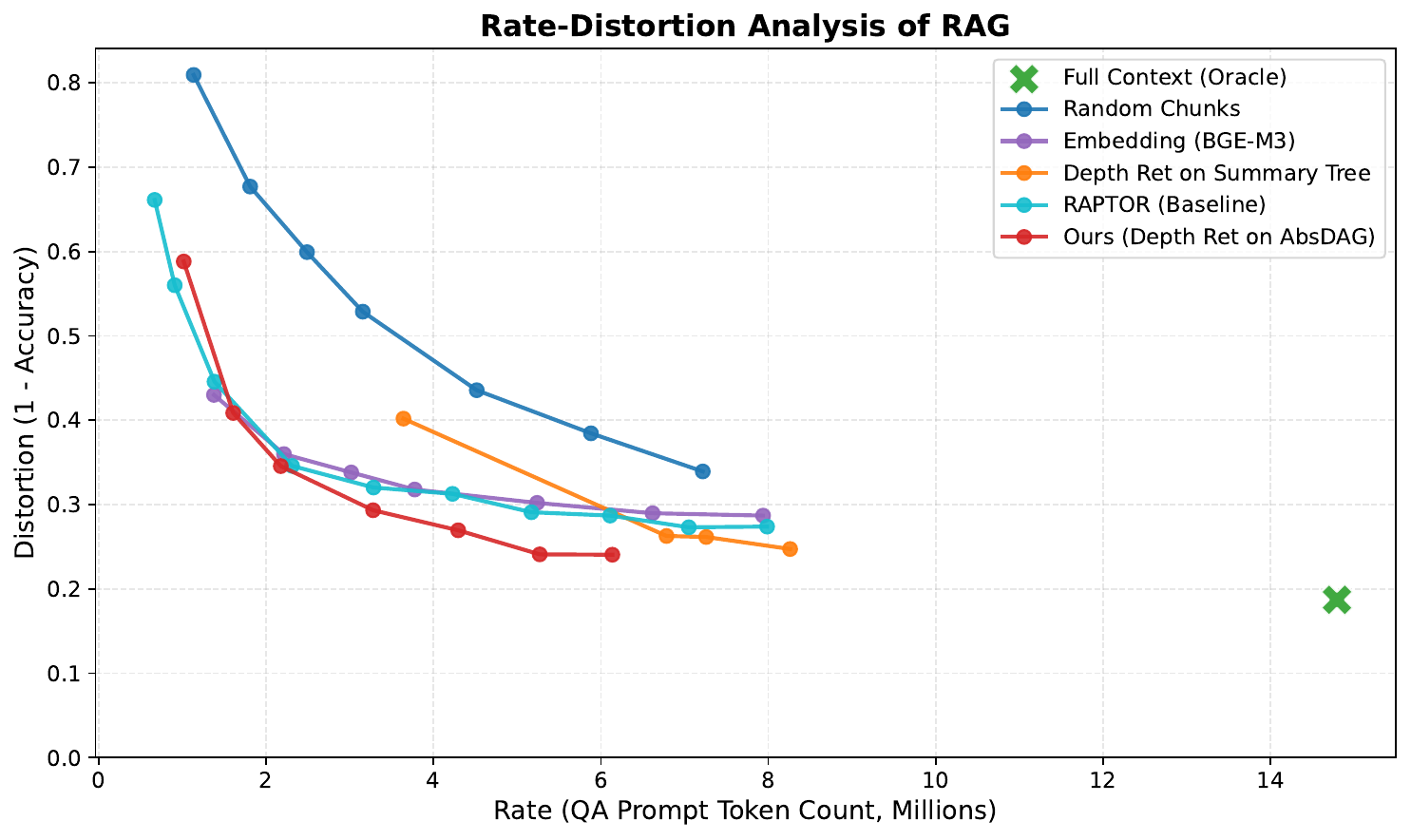}
  \caption{Rate-distortion assessment of RAG. Red denotes our RAG solution leveraging proposed semantic hierarchy.}
    \label{fig:rag}
\end{wrapfigure}

\subsection{Rate-distortion assessment for RAG}\label{sec:results_rag}

To further demonstrate the use of the proposed semantic hierarchy and assessment framework, we propose a novel hierarchical RAG solution leveraging the proposed hierarchy as guidance for retrieval. Given a user query, we first instruct the model to select the most relevant nodes in the hierarchy through a depth first retrieval. Given the selected top N nodes, we instruct the model to project the node summaries to corresponding context and concatenate the context for QA evaluation. We perform the experiments over the QuALITY dataset in comparison to baseline solutions. \par
As shown in Figure~\ref{fig:rag}, green denotes the oracle accuracy achieved by providing full context without truncation. This serves as the asymptotic target of the RAG performance. We consider a random chunking baseline (blue) where random chunks are extracted to be used as the context for QA. We perform experiments at various rates by concatenating various numbers of random chunks based on the rate budget. We also include a baseline leveraging BGE-M3~\citep{multi2024m3} embeddings (purple) where context sentences are compared against the query with respect to embedding distances, ranked accordingly, selected and concatenated based on the ranking to form the context. In this case, the variations in rates are introduced by considering the top-N most relevant sentences based on the rate budget. We also consider a hierarchical retrieval ablation (orange) where we perform uniform chunking of 250-word chunks and construct a hierarchical tree of summaries of the chunks with a bottom-up merge of factor~2. In this case, we perform our depth based retrieval on the hierarchal summary tree as the baseline hierarchy and we form contexts at varying rates by considering the top-N most relevant nodes selected by LLMs. Finally we also compare to baseline RAPTOR~\citep{raptor} (cyan) where we used leaf chunks of 100 tokens, up to 5 tree layers, a per-query context budget of 7000 tokens, and BGE-M3 embeddings for both tree construction and retrieval. For all methods, GPT-4.1-mini serves as the QA evaluation model at temperature 0. \par
Figure~\ref{fig:rag} shows the comparison between our proposed solution (red) against baselines. In general, all the approaches demonstrate consistent improvement in RAG performance with increasing rates, evidenced with monotonically decreasing rate-distortion curves. It is shown that for the embedding-based baseline (purple and blue), while it achieves a strong performance at a low rate budget, the method falls short in leveraging additional rates to achieve reasoning improvement and the curve saturates at a performance level suggesting persistent mis-comprehension. In comparison to other baselines, the baseline leveraging hierarchical summaries of the context shows curves with steeper gradients (orange and red) while our proposed solution leveraging the grounded semantic hierarchy shows the best trade-off which emphasizes its capability to efficiently take advantage of the context budget to achieve reasoning improvement.

%% file: conclusion.tex
\section{Conclusions}

In this work, we propose a novel framework for extracting semantic hierarchy to support principled scheme for assessment of generation from large models grounded in information theory. Constructing the semantic hierarchy of content provides us a systematic definition of a fidelity measure to support assessment over rate-distortion trade-off across a wide range of content generation and manipulation tasks. While in this work, we demonstrate the assessment on inspecting and comparing LLM-based content summarization, the same framework can be generalized over alternative definitions of rate and/or distortion. In future work, we propose to explore optimization of models with respect to task-specific rate-distortion trade-off.

%% file: custom.bib
@inproceedings{quality,
    title = "{Q}u{ALITY}: Question Answering with Long Input Texts, Yes!",
    author = "Pang, Richard Yuanzhe  and
      Parrish, Alicia  and
      Joshi, Nitish  and
      Nangia, Nikita  and
      Phang, Jason  and
      Chen, Angelica  and
      Padmakumar, Vishakh  and
      Ma, Johnny  and
      Thompson, Jana  and
      He, He  and
      Bowman, Samuel",
    booktitle = "NAACL",
    year = "2022",
}

@inproceedings{nagle24,
    title = "Fundamental Limits of Prompt Compression: A Rate-Distortion Framework for Black-Box Language Models",
    author = "Alliot Nagle and Adway Girish and Marco Bondaschi and Michael Gastpar and Ashok Vardhan Makkuva and Hyeji Kim",
    booktitle = "NeurIPS",
    year = "2024",
}

@inproceedings{geval,
    title = "{G}-Eval: {NLG} Evaluation using Gpt-4 with Better Human Alignment",
    author = "Liu, Yang  and
      Iter, Dan  and
      Xu, Yichong  and
      Wang, Shuohang  and
      Xu, Ruochen  and
      Zhu, Chenguang",
    year = "2023",
    booktitle = "EMNLP"
}

@inproceedings{questeval,
    title = "{Q}uest{E}val: Summarization Asks for Fact-based Evaluation",
    author = "Scialom, Thomas  and
      Dray, Paul-Alexis  and
      Lamprier, Sylvain  and
      Piwowarski, Benjamin  and
      Staiano, Jacopo  and
      Wang, Alex  and
      Gallinari, Patrick",
    booktitle = "EMNLP",
    year = "2021",
}

@inproceedings{wang20asking,
    title = "Asking and Answering Questions to Evaluate the Factual Consistency of Summaries",
    author = "Wang, Alex  and
      Cho, Kyunghyun  and
      Lewis, Mike",
    booktitle = "ACL",
    year = "2020",
}

@inproceedings{wordnet,
    title = "WordNet: A Lexical Database for English.",
    author = "George A. Miller",
    booktitle = "Communications of the ACM",
    volume ="38",
    year = "1995",
}

@book{rd,
    title = "Rate Distortion Theory: A Mathematical Basis for Data Compression",
    author = "Toby Berger",
    year = "1971",
}

@book{info,
    title = "Elements of Information Theory",
    author = "Thomas M. Cover and Joy A. Thomas",
    Edition = "2nd",
    year = "2005",
}

@article{rst,
title = {Rhetorical Structure Theory: Toward a functional theory of text organization},
author = {William C. Mann and Sandra A. Thompson},
pages = {243--281},
volume = {8},
number = {3},
journal = {Text - Interdisciplinary Journal for the Study of Discourse},
year = {1988},
}

@inproceedings{rouge,
    title = "{ROUGE}: A Package for Automatic Evaluation of Summaries",
    author = "Lin, Chin-Yew",
    booktitle = "Text Summarization Branches Out",
    month = jul,
    year = "2004",
    address = "Barcelona, Spain",
    publisher = "Association for Computational Linguistics",
    url = "https://aclanthology.org/W04-1013/",
    pages = "74--81"
}

@article{multi2024m3,
  title={M3-Embedding: Multi-linguality, multi-functionality, multi-granularity text embeddings through self-knowledge distillation},
  author={Jianlv Chen and Shitao Xiao and Peitian Zhang and Kun Luo and Defu Lian and Zheng Liu},
  journal={arXiv preprint arXiv:2402.03216},
  year={2024}
}

@inproceedings{lee25,
    title = "How Well do LLMs Compress Their Own Chain-of-Thought? A Token Complexity Approach",
    author = "Ayeong Lee and Ethan Che and Tianyi Peng",
    booktitle = "arXiv:2503.01141v1",
    year = "2025",
}

@inproceedings{young25,
    title = "Radio: Rate–Distortion Optimization for Large Language Model Compression",
    author = "Sean I. Young",
    booktitle = "ICML",
    year = "2025",
}

@article{bigbird,
  title={Big bird: Transformers for longer sequences},
  author={Zaheer, Manzil and Guruganesh, Guru and Dubey, Kumar Avinava and Ainslie, Joshua and Alberti, Chris and Ontanon, Santiago and Pham, Philip and Ravula, Anirudh and Wang, Qifan and Yang, Li and others},
  journal={NeurIPS},
  year={2020}
}

@article{longformer,
  title={Longformer: The Long-Document Transformer},
  author={Iz Beltagy and Matthew E. Peters and Arman Cohan},
  journal={arXiv:2004.05150},
  year={2020},
}

@article{category,
  title={Document Understanding, Measurement, and Manipulation Using Category Theory},
  author={Jared Claypoole and Yunye Gong and Noson S. Yanofsky and Ajay Divakaran},
  journal={arXiv:2510.21553},
  year={2025},
}

@article{bleu,
  title={BLEU: a Method for Automatic Evaluation of Machine Translation},
  author={Kishore Papineni and Salim Roukos and Todd Ward and Wei-Jing Zhu},
  journal={ACL},
  year={2002},
}

@article{rae2020,
  title={Compressive Transformers for Long-Range Sequence Modelling},
  author={Jack W. Rae and Anna Potapenko and Siddhant M. Jayakumar and Chloe Hillier and Timothy P. Lillicrap},
  journal={ICLR},
  year={2020}
}

@article{chevalier2023,
  title={Adapting Language Models to Compress Contexts},
  author={Alexis Chevalier and Alexander Wettig and Anirudh Ajith and Danqi Chen},
  journal={EMNLP},
  year={2023}
}

@inproceedings{raptor,
    title={RAPTOR: Recursive Abstractive Processing for Tree-Organized Retrieval},
    author={Sarthi, Parth and Abdullah, Salman and Tuli, Aditi and Khanna, Shubh and Goldie, Anna and Manning, Christopher D.},
    booktitle={ICLR},
    year={2024}
}

@inproceedings{graphrag,
    title={From Local to Global: A Graph RAG Approach to Query-Focused Summarization},
    author={Darren Edge and Ha Trinh and Newman Cheng and Joshua Bradley and Alex Chao and Apurva Mody and Steven Truitt and Dasha Metropolitansky and Robert Osazuwa Ness and Jonathan Larson},
    booktitle={arXiv:2404.16130},
    year={2024}
}

@inproceedings{conceptnet,
    title={ConceptNet 5.5: an open multilingual graph of general knowledge},
    author={Robyn Speer and Joshua Chin and Catherine Havasi},
    booktitle={AAAI},
    year={2017}
}
